\documentclass[10pt,twocolumn,letterpaper]{article}

\usepackage[pagenumbers]{cvpaper}
\usepackage{xcolor}
\usepackage{tabularx}
\usepackage{amsmath,amssymb,mathtools}
\usepackage{xcolor}
\usepackage{enumitem}

\usepackage{tikz}
\usetikzlibrary{arrows.meta,fit,positioning}

\providecommand{\hyvol}{\textsc{HyVoL}}
\providecommand{\gram}{\textsc{Gram}}
\providecommand{\hypergram}{\textsc{HyperGram}}
\providecommand{\trianglemodel}{\textsc{Triangle}}
\providecommand{\pmrl}{\textsc{PMRL}}
\providecommand{\vast}{\textsc{VAST}}

\setlist[itemize]{
    leftmargin=*,
    topsep=2pt,
    itemsep=1pt,
    parsep=0pt
}
\usepackage{amsmath,amssymb}
\usepackage{multirow}
\usepackage{booktabs}
\PassOptionsToPackage{table}{xcolor}
\usepackage{xcolor}
\usepackage{graphicx}

\IfFileExists{dsfont.sty}{\usepackage{dsfont}}{%
  \IfFileExists{bbm.sty}{\usepackage{bbm}}{%
  }}

\providecommand{\hyvol}{HyVoL}

\definecolor{cvblue}{rgb}{0.21,0.49,0.74}
\usepackage[breaklinks,colorlinks,allcolors=cvblue]{hyperref}
\usepackage{booktabs}
\usepackage{multirow}
\usepackage{graphicx}
\usepackage[table]{xcolor}
\renewcommand{\hyvol}{\textsc{HyVoL}\xspace}

\title{Hypergraph-Regularized Gramian Volumes for Multimodal Retrieval}

\author{Anindya Nag \quad Ambuj Mehrish \quad Sebastiano Vascon\\
Department of Environmental Science, Informatics and Statistics\\
Ca' Foscari University of Venice, Italy\\
{\tt\small \{anindya.nag, ambuj.mehrish, sebastiano.vascon\}@unive.it}
}
\hypersetup{
  pdftitle={Hypergraph-Regularized Gramian Volumes for Multimodal Retrieval},
  pdfauthor={Anindya Nag, Ambuj Mehrish, Sebastiano Vascon}
}

\begin{document}
\maketitle
\begin{abstract}
Volume-based multimodal retrieval jointly scores a text query with a candidate's video, audio, and subtitle embeddings. While this approach captures higher-order within-candidate alignment, the score remains candidate-local, and semantically related training samples primarily serve as contrastive negatives. This work introduces Hypergraph-Regularized Gramian Volumes (HyVol), a training-time module that incorporates these semantic relations prior to evaluating the original volume loss. Document hyperedges connect the observed modalities of each candidate, whereas semantic hyperedges link candidates whose detached captions are mutual top-k neighbors. A shallow gated hypergraph network applies residual corrections to the modality embeddings. Presence masks exclude unavailable streams from message passing, and identity padding preserves the determinant of the observed Gram submatrix without feature imputation. As refinement operates on embeddings rather than scores, the same construction applies to both Gram and HyperGram. We remove the hypergraph after training, leaving the backbone-only architecture, original scoring function, and retrieval cost unchanged. We train both backbones on a 150K-clip subset of VAST-27M and evaluate zero-shot performance on six benchmarks. Under the paired protocol, HyVol improves R@1 across all five retrieval benchmarks, with video-to-text gains reaching $+8.3$ on MSR-VTT and $+7.6$ on VATEX. Under missing-modality masking, the V2T margin remains positive in all experimental settings, although the T2V margin becomes slightly negative in four.
\end{abstract} 
\section{Introduction}
\label{sec:intro}

Multimodal retrieval encompasses more than matching text to visual frames; modalities such as sound, speech, and subtitles can each contribute evidence for a caption. Models including CLIP~\cite{Radford2021LearningTV} and ALIGN~\cite{jia2021align} learn shared embedding spaces via pairwise similarity. In contrast, multimodal extensions align multiple streams through an anchor or integrate them before scoring~\cite{Girdhar2023ImageBindOE,liu2024valor,chen2023vast,Zhu2023LanguageBindEV}. These advancements have enabled broad transfer and inspired objectives that jointly score multiple embeddings.

Gramian alignment represents one such approach. Given a query and a candidate's modality embeddings, \gram{} computes the volume of the parallelotope they span as an alignment score~\cite{cicchetti2025gramian}. For two unit vectors, this volume is $\sqrt{1-\cos^2\theta}=|\sin\theta|$; for larger sets, it characterizes their joint linear configuration. Related methods utilize triangle area~\cite{cicchetti2026triangle}, encourage a dominant singular direction~\cite{liu2026principled}, or integrate Euclidean and Lorentz-hyperbolic volumes~\cite{na2026hypergram}. Additional research explores volume-based regularization~\cite{you2025mover}, higher-order contrastive fusion~\cite{koutoupis2026more}, and hyperbolic vision--language representations~\cite{desai2023hyperbolic,pal2025compositional}. Despite geometric distinctions, these retrieval scores remain candidate-local: the score for candidate $j$ depends solely on the query and the modalities of candidate $j$.

This candidate-local structure fails to model potentially valuable relationships among documents. While other clips are included in the contrastive objective as negatives, semantically related examples, such as alternative views of a concert or kayaking scene, do not explicitly inform each other's representations. This limitation is especially significant when modalities are incomplete~\cite{ma2021smil,ma2022multimodal}, as each example must be learned and evaluated from a reduced set of observed streams without leveraging relationships to semantically similar training instances.

Hypergraph-Regularized Gramian Volumes (\hyvol{}) address this limitation by serving as a training-time module that propagates information among documents before computing the volume loss. The hypergraph consists of two edge types. A document edge links the observed modalities within a single clip, while absent streams remain unconnected. A semantic edge links clips whose detached caption features are mutual top-$k$ neighbors. A shallow gated hypergraph network performs vertex-to-edge-to-vertex propagation, generating a residual correction to the modality embeddings. The original volume objective is then evaluated using these corrected embeddings.

\hyvol{} modifies embeddings rather than the scoring function, enabling a unified approach to regularize both the Euclidean volume of \gram{} and the hybrid Lorentzian volume of \hypergram{}. The hypergraph is utilized exclusively during training and is omitted at test time. Consequently, the deployed system consists solely of the backbone, introduces no additional retrieval cost or parameters, and prevents gallery embeddings from accessing the test query. Missing streams are addressed without feature imputation; they are disconnected from the graph and represented in the Gram matrix by orthonormal phantom axes, thereby preserving the determinant of the observed-modality submatrix.

\hyvol{} is instantiated on \gram{}~\cite{cicchetti2025gramian} and \hypergram{}~\cite{na2026hypergram} using identical encoders, datasets, optimization procedures, batch sizes, and training schedules for each paired comparison. Across six zero-shot benchmarks, \hyvol{} improves performance on five retrieval tasks under the specified protocol while maintaining comparable results on the audio-classification probe. On MSR-VTT the improvement grows with modality arity, though this trend is benchmark-dependent. On the four-modality MSR-VTT~\cite{MSRVTT} benchmark, text-to-video R@1 increases from $51.8$ to $55.8$ for \gram{} and from $52.3$ to $56.1$ for \hypergram{}. The corresponding improvements on VATEX~\cite{wang2019vatex} are $5.6$ and $3.5$ points, respectively, while the video-to-text improvement reaches $+8.3$ on MSR-VTT. 

Our contributions are threefold. First, we examine the candidate-local structure of volume-based multimodal alignment under incomplete modality sets. Second, we introduce \hyvol{}, a training-time module combining presence-masked document edges, mutual top-\(k\) semantic edges, and gated hypergraph refinement. We handle missing streams without imputation through disconnected vertices and orthonormal phantom axes, supporting both Euclidean and Lorentzian volumes; the module is removed after training and adds no inference cost. Third, we evaluate \hyvol{} with two volume-based backbones across six zero-shot benchmarks under a unified protocol, with sensitivity analyses on the semantic neighbourhood size and on gallery-side modality loss.

% The main contributions are as follows:

% \begin{itemize}
% \itemsep2pt
% \item We identify the candidate-local structure of volume-based multimodal alignment and study its implications under incomplete modality sets.
% \item We introduce \hyvol{}, a training-time module that integrates presence-masked document edges, mutual top-$k$ semantic edges, and gated hypergraph refinement. We address missing data streams without imputation by using disconnected graph vertices and orthonormal phantom axes, a construction suitable for both Euclidean and Lorentzian volumes. The module is removed post-training and does not affect inference.
% \item We evaluate \hyvol{} using two volume-based backbone architectures and six zero-shot benchmarks within a unified protocol, including ablation studies on neighborhood size and hypergraph components.
% \end{itemize}     
\section{Related Work}
\label{sec:related}

\noindent\textbf{Contrastive Vision--Language and Multimodal Foundations.}
CLIP~\cite{Radford2021LearningTV} and ALIGN~\cite{jia2021align} introduced contrastive retrieval within a unified image--text embedding space. Subsequent studies have explored sigmoid-based objectives~\cite{Zhai2023SigmoidLF}, open and scalable backbone architectures~\cite{ilharco_gabriel_2021_5143773,Sun2023EVACLIPIT}, and comprehensive analyses of contrastive learning~\cite{Uesaka2024UnderstandingMC}. This foundational framework has been extended to video domains via frame-level pooling~\cite{Luo2021CLIP4ClipAE,gabeur2020multi,liu2022umt}, captioner-based transfer~\cite{yan2022videococa}, and large-scale video--language models~\cite{Zeng2023TVTSv2LO,Ye2022HiTeAHT,Wang2022InternVideoGV,Wang2024InternVideo2SV,Xu2023mPLUG2AM,Wang2022OmniVLOF,Zhao2024VideoPrismAF}. Parallel advancements include audio--text~\cite{CLAP2022,oncescu2021audio,Nagrani2022LearningAM,Zhao2021ConnectingTD} and point--text retrieval~\cite{Zhang2021PointCLIPPC}. Omni-modal models such as \vast{}~ \cite{chen2023vast} and VALOR~\cite{liu2024valor} integrate video, audio, and subtitle modalities. ImageBind~\cite{Girdhar2023ImageBindOE}, CLIP4VLA~\cite{Ruan2023AccommodatingAM}, LanguageBind~\cite{Zhu2023LanguageBindEV}, OmniBind~\cite{zhang2025omnibind}, and UniBind~\cite{lyu2024unibind} achieve modality alignment through image, audio, text, or learned anchors, occasionally utilizing cross-modal fusion~\cite{Li2021AlignBF}. While most approaches optimize alignment using an anchor or fused representation, explicit modeling of agreement among constituent streams is often lacking. Incorporating complementary modalities can facilitate query disambiguation~\cite{Yoon2023HEARHE}.

\noindent\textbf{Geometric Higher-Order Alignment.}
Recent methods have replaced pairwise cosine similarity with joint geometric objectives. \gram{}~\cite{cicchetti2025gramian} leverages the Gramian volume defined by the query and modality vectors, whereas \trianglemodel{}~\cite{cicchetti2026triangle} utilizes the area of the triangle they form. PMRL~\cite{liu2026principled} encourages rank-one alignment through the dominant singular value, and Symile~\cite{saporta2024symile} maximizes a total-correlation bound across multiple streams. \hypergram{}~\cite{na2026hypergram} generalizes the Gramian volume to a Lorentz hyperboloid, thereby linking multimodal alignment with hyperbolic representation learning~\cite{nickel2017poincare,ganea2018hyperbolic} and hierarchical vision--language embeddings~\cite{desai2023hyperbolic,pal2025compositional}. Although these objectives differ in geometric formulation, they treat modalities symmetrically and evaluate each document using its own embeddings. As a result, inter-document relations are primarily introduced through contrastive negatives, rather than being directly encoded in the learned representations.

\noindent\textbf{Hypergraph and Relational Representation Learning.}
Hypergraphs represent higher-order relationships by allowing a single edge to connect multiple vertices. Hypergraph neural networks~\cite{feng2019hypergraph,gao2023hgnnp}, including dynamic variants~\cite{jiang2019dynamic}, propagate information through vertex--edge--vertex incidence structures. In contrast, graph diffusion and re-ranking techniques refine retrieval over the gallery manifold~\cite{iscen2017efficient,zhong2017reranking,geigle2022retrieve}, typically operating on a fixed candidate set during inference. Approaches to incomplete-modality learning include reconstruction, prompting, and modality dropout~\cite{ma2021smil,ma2022multimodal,lee2023multimodal,wu2024deep}. \hyvol{} utilizes a hypergraph solely during training to regularize embeddings for a volume-based objective and evaluates observed modalities without imputation. The hypergraph is omitted at test time, ensuring that the deployed backbone and retrieval cost remain unchanged. Section~\ref{sec:method} provides further details on edge construction and gated refinement.   
\section{Methodology}
\label{sec:method}
\subsection{Problem Formulation}
\label{sec:problem}

Consider a training mini-batch containing \(B\) matched caption--document pairs, indexed by \(i\in\{1,\ldots,B\}\). During retrieval, caption \(i\) is compared with each candidate document \(j\in\{1,\ldots,B\}\); the pair where \(j=i\) is considered positive, while \(j\neq i\) corresponds to an in-batch negative. The text encoder maps caption \(i\) to an \(\ell_2\)-normalised embedding \(\mathbf{t}_i\in\mathbb{R}^{D}\).
\begin{figure*}[h]
    \centering
    \includegraphics[width=0.75\textwidth]{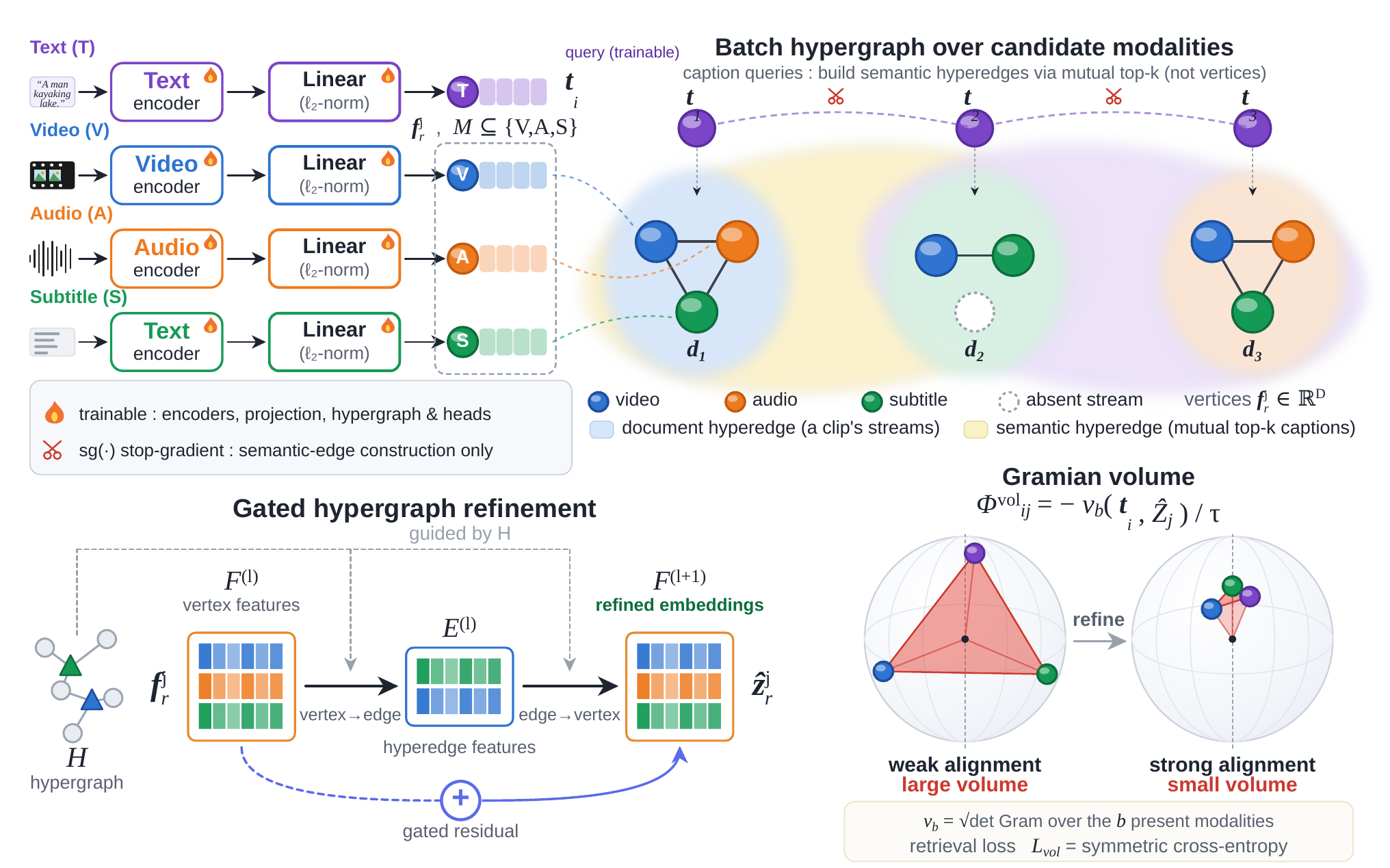}
    \caption{\textbf{Overview of \hyvol.} Available modality embeddings form document hyperedges, while detached captions define mutual top-\(k\) semantic hyperedges across documents. Gated message passing refines the embeddings before \gram{}/\hypergram{} volume scoring; the hypergraph is removed at inference.}
    \label{fig:placeholder}
\end{figure*}
A candidate document \(j\) may contain a subset \(\mathcal{O}_j\subseteq\mathcal{M}\) of the supported document modalities \(\mathcal{M}\), such as video, audio, and subtitles. For each available modality \(r\in\mathcal{O}_j\), the corresponding encoder produces an \(\ell_2\)-normalised embedding
\(\mathbf{f}^{\,j}_{r}\in\mathbb{R}^{D}\). We define the modality-availability indicator as: $p_{j,r}:= \mathbf{1}\!\left[r\in\mathcal{O}_j\right]$.

The \gram{}~\cite{cicchetti2025gramian} and \hypergram{}~\cite{na2026hypergram} frameworks assign an alignment volume to the caption embedding and the set of available candidate-modality
embeddings. \gram{} uses a Euclidean Gramian volume, whereas \hypergram{} combines Euclidean and Lorentz-hyperbolic Gramian volumes. Let \(\mathcal{G}_{\mathrm{vol}} := \{\textsc{Gram},\textsc{HyperGram}\}\)
denote the set of volume backbones. For a caption--candidate pair \((i,j)\), this work
defines
\begin{equation}
    v_{ij}^{(b)}
    :=
    \nu_b\!\left(
        \mathbf{t}_i,
        \left\{
            \mathbf{f}^{\,j}_{r}
            : r\in\mathcal{O}_j
        \right\}
    \right),
    \qquad b\in\mathcal{G}_{\mathrm{vol}},
    \label{eq:base_volume}
\end{equation}
where \(\nu_b\) denotes the volume function associated with backbone \(b\), and \(v_{ij}^{(b)}\) is the scalar volume assigned to caption \(i\) and candidate \(j\). We refer the reader to the original
papers~\cite{cicchetti2025gramian,na2026hypergram} for the corresponding
Gram-matrix and Lorentzian constructions.

In both backbones, a smaller volume indicates stronger alignment. The volume is therefore converted into a contrastive retrieval logit by reversing its ordering and applying a positive temperature:
\begin{equation}
z^{(b)}_{ij} := -\frac{v^{(b)}_{ij}}{\tau}, \qquad \tau > 0,
\label{eq:logit}
\end{equation}
where \(\tau>0\) is a learned temperature. A single \(\tau\) is used
throughout: each \hyvol{} model is trained on one backbone, and the same
scalar scales every retrieval logit in Sec.~\ref{sec:objective}. A smaller volume therefore yields a larger logit, encouraging the matched candidate \(j=i\) to rank above in-batch negatives \(j\neq i\). \hyvol{} leaves the backbone volume \(\nu_b\) unchanged and instead uses batch-level relations to refine the available candidate-modality embeddings before computing the original volume and retrieval logit.

\subsection{Hypergraph Construction}
\label{sec:graph}
For a mini-batch containing \(B\) documents, one vertex row is defined for each candidate and modality pair.
\(\mathcal{V}=\{(j,r):j\in[B],\,r\in\mathcal{M}\}\).
A row \((j,r)\) is considered active with feature \(\mathbf{f}^{\,j}_{r}\) when \(p_{j,r}=1\); otherwise, both its feature and incidence are set to zero. Caption embeddings are excluded from the graph and are used solely, with stopped gradients, to identify semantic relations among documents. This approach prevents refined candidate embeddings from directly incorporating their paired captions and maintains query independence during inference. The graph consists of both document and semantic hyperedges.

\noindent\textbf{Document hyperedges.}
For each document \(j\), one hyperedge connects all of its available modality vertices. Its incidence matrix is
\begin{equation}
    [H_{\mathrm{doc}}]_{(q,r),j}
    =
    \begin{cases}
        1, & q=j \text{ and } p_{j,r}=1,\\
        0, & \text{otherwise}.
    \end{cases}
    \label{eq:hdoc}
\end{equation}
here, \(H_{\mathrm{doc}}\in\{0,1\}^{|\mathcal{V}|\times B}\) is the
document-hyperedge incidence matrix: row \((q,r)\) represents modality
\(r\) of document \(q\), and column \(j\) represents document \(j\).
Nonzero entries select the available modalities, allowing hyperedges
to have different cardinalities.

\noindent\textbf{Semantic hyperedges.}
Document hyperedges exchange information only within a candidate. To also connect related candidates, we construct semantic neighbourhoods from the paired training captions. Let
\begin{equation}
    T=
    \begin{bmatrix}
        \mathbf{t}_1^{\top}\\[-2pt]
        \vdots\\[-2pt]
        \mathbf{t}_B^{\top}
    \end{bmatrix}
    \in\mathbb{R}^{B\times D},
    \qquad
    C=TT^{\top},
    \label{eq:caption_similarity}
\end{equation}
where \(C_{ij}\) is the cosine similarity between captions \(i\) and \(j\). The diagonal of \(C\) is ignored during neighbour selection. Define
\begin{equation}
    S_{ij}
    =\mathbf{1}\!\left[j\in\operatorname{top}\text{-}k(C_{i,:})\right],
    \qquad
    N_{ij}=S_{ij}S_{ji}.
    \label{eq:mutual}
\end{equation}
Hence, \(N_{ij}=1\) only when \(i\) and \(j\) select each other as neighbours. The mutual criterion reduces one-sided connections. We additionally apply symmetric edge dropout to the retained set during
training, so that refinement does not come to depend on a single neighbour. We learn the relative importance of the retained neighbours from the detached caption embeddings. Let
\begin{equation}
    \mathbf{x}_i=W_a\mathbf{t}_i,
    \qquad
    e_{ij}
    =\operatorname{LeakyReLU}_{0.2}\!\left(
        \mathbf{a}^{\top}[\mathbf{x}_i\,\|\,\mathbf{x}_j]
    \right),
    \label{eq:attention_score}
\end{equation}
where \([\cdot\,\|\,\cdot]\) denotes concatenation. For retained edges, the normalised attention and its symmetric form are
\begin{align}
    A_{ij}
    &=
    \frac{\exp(e_{ij})}
    {\sum_{m:N_{im}=1}\exp(e_{im})},
    \qquad \text{if } N_{ij}=1,
    \label{eq:attention}\\
    \widetilde{A}
    &=\tfrac{1}{2}\left(A+A^{\top}\right),
    \label{eq:symmetric_attention}
\end{align}
with \(A_{ij}=0\) when \(N_{ij}=0\), and isolated rows set to zero. Eq.~\eqref{eq:attention} is row stochastic and therefore asymmetric, whereas the mutual criterion of Eq.~\eqref{eq:mutual} is undirected; symmetrising gives each retained pair a single magnitude in Eq.~\eqref{eq:hsem}.

We create one semantic hyperedge for each document \(i\). It contains the available modalities of \(i\) and those of its mutual neighbours, weighted by
\(\widetilde{A}\):
\begin{equation}
    [H_{\mathrm{sem}}]_{(j,r),i}
    =p_{j,r}\left(\mathbf{1}[j=i]+\widetilde{A}_{ji}\right),
    \label{eq:hsem}
\end{equation}
The complete weighted incidence matrix is the column-wise concatenation $H=\left[H_{\mathrm{doc}}\,\|\,H_{\mathrm{sem}}\right]$.
\subsection{Gated Hypergraph Refinement}
\label{sec:refine}

Let \(F^{(0)}\in\mathbb{R}^{|\mathcal{V}|\times D}\) contain the initial vertex features, with inactive rows set to zero. \hyvol{} alternates vertex-to-hyperedge and hyperedge-to-vertex aggregation. Define
\(\Delta_E=\operatorname{diag}(H^\top\mathbf{1})\) and
\(\Delta_V=\operatorname{diag}(H\mathbf{1})\), with degrees clamped below at \(1\). Layer \(\ell\) computes
\begin{align}
    E^{(\ell)}
    &=\operatorname{GELU}\!\left(
        \Delta_E^{-1}H^\top F^{(\ell)}W_V^{(\ell)}
    \right),
    \label{eq:v2e}\\
    M^{(\ell)}
    &=\Delta_V^{-1}H E^{(\ell)}W_E^{(\ell)},
    \label{eq:e2v_message}\\
    F^{(\ell+1)}
    &=F^{(\ell)}
      +\tanh(g_\ell)\,\phi_\ell\!\left(M^{(\ell)}\right),
    \label{eq:e2v}
\end{align}
Here, \(W_V^{(\ell)}\) and \(W_E^{(\ell)}\) are learned projections, while \(g_\ell\) is a learned scalar that gates the residual correction. We use GELU for \(\phi_\ell\) in intermediate layers and the identity in the final layer, with at most two layers to limit over-smoothing.

For each available modality, let
\(\mathbf{u}^{\,j}_{r}=F^{(L)}[(j,r)]\). The backbone receives the normalized feature
\(\widehat{\mathbf{z}}^{\,j}_{r}
=\mathbf{u}^{\,j}_{r}/\|\mathbf{u}^{\,j}_{r}\|_2\).
For the auxiliary alignment loss, we pool the refined modalities as
\begin{equation}
    \mathbf{h}_j
    =\operatorname{norm}\!\left(
        W_{\mathrm{pool}}
        \frac{\sum_{r\in\mathcal{M}}
        p_{j,r}\mathbf{u}^{\,j}_{r}}
        {\sum_{r\in\mathcal{M}}p_{j,r}}
    \right),
    \label{eq:docemb}
\end{equation}
The availability mask excludes missing streams from the pool. 

\subsection{Training Objective}
\label{sec:objective}
For document \(j\), let $\widehat{Z}^{\,j} =\{\widehat{\mathbf{z}}^{\,j}_{r}:r\in\mathcal{O}_j\}$ be its available refined embeddings. The backbone volume logits are $   \Phi^{\mathrm{vol}}_{ij} =-\frac{\nu_b(\mathbf{t}_i,\widehat{Z}^{\,j})}{\tau}$, where \(\tau\) is a learned temperature of  Eq.~\eqref{eq:logit}. The
auxiliary document term in Eq.~\eqref{eq:document_logits} reuses the same
\(\tau\) rather than introducing a second one. The two logit families are not on identical scales, since \(\Phi^{\mathrm{vol}}\) is formed from volumes at the document's full arity whereas \(\Phi^{\mathrm{doc}}\) uses the arity-two volume \(\nu_{b,2}\); we do not tune a separate temperature for the auxiliary term and instead let \(w_{\mathrm{doc}}\) set its relative weight.  The primary retrieval loss is the bidirectional cross-entropy
\begin{equation}
    \mathcal{L}_{\mathrm{vol}}
    =\frac{1}{2}\left[
        \operatorname{CE}(\Phi^{\mathrm{vol}},\mathbf{y})
        +
        \operatorname{CE}((\Phi^{\mathrm{vol}})^{\top},\mathbf{y})
    \right],
    \label{eq:area}
\end{equation}
where \(\mathbf{y}\) identifies the matching caption--document pairs in the gathered batch.

Two auxiliary losses are applied to the hypergraph output. First, we align each caption with the pooled document embedding in Eq.~\eqref{eq:docemb}. Let \(\nu_{b,2}\) denote the arity-two form of the same backbone volume. We define
\begin{equation}
    \Phi^{\mathrm{doc}}_{ij}
    =-\frac{\nu_{b,2}(\mathbf{t}_i,\mathbf{h}_j)}{\tau},
    \label{eq:document_logits}
\end{equation}
and compute \(\mathcal{L}_{\mathrm{doc}}\) using the same bidirectional cross-entropy as Eq.~\eqref{eq:area}.

Second, a variance regulariser discourages the refined features from collapsing. Let \(U_r\) stack the pre-normalisation features \(\mathbf{u}^{\,j}_{r}\) for the available samples of modality \(r\), and let \(U_{\mathrm{doc}}\) stack the document features \(\mathbf{h}_j\). With \(\sigma(X)=\sqrt{\operatorname{Var}(X)+\epsilon}\) denoting the vector of feature-wise standard deviations, we use
\begin{equation}
    \begin{aligned}
        \mathcal{L}_{\mathrm{reg}}
        &=
        \sum_{r\in\mathcal{M}}
        \operatorname{mean}\!\left[
            \operatorname{ReLU}\!\left(1-\sigma(U_r)\right)
        \right]
        \\
        &\quad+
        \operatorname{mean}\!\left[
            \operatorname{ReLU}\!\left(1-\sigma(U_{\mathrm{doc}})\right)
        \right],
    \end{aligned}
    \label{eq:vicreg}
\end{equation}

The complete objective is
\begin{equation}
    \mathcal{L}
    =\mathcal{L}_{\mathrm{vol}}
    +w_{\mathrm{doc}}\mathcal{L}_{\mathrm{doc}}
    +w_{\mathrm{reg}}\mathcal{L}_{\mathrm{reg}}
    +w_{\mathrm{itm}}\mathcal{L}_{\mathrm{itm}},
    \label{eq:total}
\end{equation}
 Here, \(\mathcal{L}_{\mathrm{itm}}\) denotes the backbone's existing caption--candidate matching loss, which is retained without modification and weighted by $w_{\mathrm{itm}}$. This loss employs binary cross-entropy on a fused caption--candidate matching head, utilizing each batch row's positive pair and in-batch negatives. Consequently, the \hyvol{} objective differs from its paired backbone solely through \(\mathcal{L}_{\mathrm{doc}}\) and \(\mathcal{L}_{\mathrm{reg}}\); no additional matching head, reconstruction loss, or link-prediction objective is introduced. Since refinement occurs prior to \(\nu_b\), the same module is applicable to both \gram{}~\cite{cicchetti2025gramian} and \hypergram{}~\cite{na2026hypergram}.

\subsection{Training, Inference, and Missing Modalities}
\label{sec:traintest}
\noindent\textbf{Training and inference.}
During training, \hyvol{} refines candidate embeddings prior to evaluating Eq.~\eqref{eq:total}, while caption embeddings remain fixed. During inference, the module is omitted, and the trained encoders utilize the original \gram{}~\cite{cicchetti2025gramian} or \hypergram{}~\cite{na2026hypergram} score. Consequently, retrieval does not require hypergraph construction or message passing.

\noindent\textbf{Missing modalities.}
The availability indicator \(p_{j,r}\) of Sec.~\ref{sec:problem} excludes
missing modalities from the incidence matrices, the pooled document
representation, and the refined set \(\widehat{Z}^{\,j}\). Absent streams are neither imputed nor propagated. To maintain fixed-size Gram matrices for batched computation, the corresponding rows and columns are numerically replaced with identity entries. Up to a simultaneous row and column permutation,
\begin{equation}
    \begin{aligned}
        G_{\mathrm{masked}}
        &=\operatorname{diag}(G_{\mathrm{present}},I),\\
        \det(G_{\mathrm{masked}})
        &=\det(G_{\mathrm{present}}).
    \end{aligned}
    \label{eq:masked_gram}
\end{equation}
Thus, identity padding contributes a factor of one and recovers the volume computed from the observed modalities only. We apply the same rule to the Euclidean Gram matrix of \gram{} and the corresponding Lorentzian matrix of \hypergram{}.    \section{Experiments}
\label{sec:experiments}

\subsection{Experimental Setup}
\label{sec:setup}
\begin{table*}[t]
\small
\centering
\caption{
Zero-shot retrieval results on MSR-VTT~\cite{MSRVTT}.
\hyvol is applied on top of \gram{}~\cite{cicchetti2025gramian} and \hypergram{}~\cite{na2026hypergram}.
\textsuperscript{\S} published numbers (reference only);
\textsuperscript{$\star$} authors' released checkpoint, evaluated on our protocol;
\textsuperscript{$\dag$} trained from the authors' released code at their recipe.
}
\label{tab:msrvtt_results}
\footnotesize
\setlength{\tabcolsep}{3pt}
\renewcommand{\arraystretch}{1.05}
\resizebox{1.7\columnwidth}{!}{%
\begin{tabularx}{\textwidth}{
    @{}
    l
    *{12}{>{\centering\arraybackslash}X}
    @{}
}
\toprule
\multirow{3}{*}{Method}
& \multicolumn{4}{c}{\textbf{T--V}}
& \multicolumn{4}{c}{\textbf{T--VA}}
& \multicolumn{4}{c}{\textbf{T--VAS}} \\
\cmidrule(lr){2-5}
\cmidrule(lr){6-9}
\cmidrule(lr){10-13}

& \multicolumn{2}{c}{T2V}
& \multicolumn{2}{c}{V2T}
& \multicolumn{2}{c}{T2V}
& \multicolumn{2}{c}{V2T}
& \multicolumn{2}{c}{T2V}
& \multicolumn{2}{c}{V2T} \\
\cmidrule(lr){2-3}
\cmidrule(lr){4-5}
\cmidrule(lr){6-7}
\cmidrule(lr){8-9}
\cmidrule(lr){10-11}
\cmidrule(lr){12-13}

& R@1 & R@10 & R@1 & R@10
& R@1 & R@10 & R@1 & R@10
& R@1 & R@10 & R@1 & R@10 \\
\midrule

Norton~\cite{lin2024norton}\textsuperscript{\S}
& 10.7 & 31.6 & - & -
& - & - & - & -
& - & - & - & - \\

UMT~\cite{Li2023UnmaskedTT}\textsuperscript{\S}
& 33.3 & 66.7 & 33.3 & 66.7
& - & - & - & -
& - & - & - & - \\

VideoCoCa~\cite{yan2022videococa}\textsuperscript{\S}
& 34.3 & 67.0 & 34.3 & 67.0
& - & - & - & -
& - & - & - & - \\

HiTeA~\cite{Ye2022HiTeAHT}\textsuperscript{\S}
& 34.4 & 69.9 & 46.8 & -
& - & - & - & -
& - & - & - & - \\

ImageBind~\cite{Girdhar2023ImageBindOE}\textsuperscript{\S}
& 36.8 & 70.0 & 36.8 & 70.0
& - & - & - & -
& - & - & - & - \\

TVTSv2~\cite{Zeng2023TVTSv2LO}\textsuperscript{\S}
& 38.2 & 73.2 & 38.2 & 73.2
& - & - & - & -
& - & - & - & - \\

UMT-L~\cite{Li2023UnmaskedTT}\textsuperscript{\S}
& 40.7 & 71.8 & 40.7 & -
& - & - & - & -
& - & - & - & - \\

InternVideo-L~\cite{Wang2022InternVideoGV}\textsuperscript{\S}
& 40.7 & - & 39.6 & -
& - & - & - & -
& - & - & - & - \\

OmniVL~\cite{Wang2022OmniVLOF}\textsuperscript{\S}
& 42.0 & 73.0 & 34.6 & 66.6
& - & - & - & -
& - & - & - & - \\

ViCLIP~\cite{Wang2023InternVidAL}\textsuperscript{\S}
& 42.4 & - & 41.3 & -
& - & - & - & -
& - & - & - & - \\

LanguageBind~\cite{Zhu2023LanguageBindEV}\textsuperscript{\S}
& 44.8 & 78.7 & 40.9 & 75.7
& - & - & - & -
& - & - & - & - \\

mPLUG-2~\cite{Xu2023mPLUG2AM}\textsuperscript{\S}
& 47.1 & 79.0 & - & -
& - & - & - & -
& - & - & - & - \\

VideoPrism-b~\cite{Zhao2024VideoPrismAF}\textsuperscript{\S}
& 51.4 & - & 50.2 & -
& - & - & - & -
& - & - & - & - \\

\vast{}~\cite{chen2023vast}\textsuperscript{\S}
& - & - & - & -
& 49.3 & 80.0 & 43.7 & 77.1
& 50.7 & 74.4 & 49.0 & 76.2 \\

\addlinespace[2pt]

\trianglemodel{}~\cite{cicchetti2026triangle}\textsuperscript{$\star$}
& - & - & - & -
& 50.5 & 79.4 & 49.7 & 80.3
& - & - & - & - \\

\pmrl{}~\cite{liu2026principled}\textsuperscript{$\star$}
& 51.3 & 81.3 & 49.6 & 81.1
& 50.9 & 81.1 & 51.0 & 80.3
& 54.3 & 79.5 & 52.9 & 79.4 \\

\gram{}~\cite{cicchetti2025gramian}\textsuperscript{$\star$}
& 51.9 & 82.4 & 48.7 & 80.7
& 52.0 & 82.0 & 49.3 & 80.7
& 51.8 & 81.9 & 50.1 & 80.3 \\

\rowcolor{gray!15}
-\quad w \hyvol
& 52.6 \textbf{\textcolor{green!50!black}{(+0.7)}}
& 82.9 \textbf{\textcolor{green!50!black}{(+0.5)}}
& 54.5 \textbf{\textcolor{green!50!black}{(+5.8)}}
& 82.8 \textbf{\textcolor{green!50!black}{(+2.1)}}
& 54.0 \textbf{\textcolor{green!50!black}{(+2.0)}}
& 83.3 \textbf{\textcolor{green!50!black}{(+1.3)}}
& 55.8 \textbf{\textcolor{green!50!black}{(+6.5)}}
& 82.2 \textbf{\textcolor{green!50!black}{(+1.5)}}
& 55.8 \textbf{\textcolor{green!50!black}{(+4.0)}}
& 83.9 \textbf{\textcolor{green!50!black}{(+2.0)}}
& 58.1 \textbf{\textcolor{green!50!black}{(+8.0)}}
& 83.7 \textbf{\textcolor{green!50!black}{(+3.4)}} \\

\hypergram{}~\cite{na2026hypergram}\textsuperscript{$\dag$}
& 50.8 & 82.0 & 48.7 & 82.9
& 51.7 & 81.7 & 49.8 & 82.1
& 52.3 & 83.0 & 50.9 & 83.2 \\

\rowcolor{gray!15}
\quad w \hyvol
& 52.5 \textbf{\textcolor{green!50!black}{(+1.7)}}
& 82.9 \textbf{\textcolor{green!50!black}{(+0.9)}}
& 54.2 \textbf{\textcolor{green!50!black}{(+5.5)}}
& 83.9 \textbf{\textcolor{green!50!black}{(+1.0)}}
& 54.2 \textbf{\textcolor{green!50!black}{(+2.5)}}
& 83.5 \textbf{\textcolor{green!50!black}{(+1.8)}}
& 56.6 \textbf{\textcolor{green!50!black}{(+6.8)}}
& 83.9 \textbf{\textcolor{green!50!black}{(+1.8)}}
& 56.1 \textbf{\textcolor{green!50!black}{(+3.8)}}
& 83.8 \textbf{\textcolor{green!50!black}{(+0.8)}}
& 59.2 \textbf{\textcolor{green!50!black}{(+8.3)}}
& 84.7 \textbf{\textcolor{green!50!black}{(+1.5)}} \\

\bottomrule
\end{tabularx}}
\end{table*}
\begin{table*}[t]
\small
\centering
\caption{
Zero-shot retrieval results on DiDeMo~\cite{DIDEMO}, ActivityNet~\cite{ACTIVITYNET}, and VATEX~\cite{wang2019vatex}.
\hyvol is applied on top of \gram{}~\cite{cicchetti2025gramian} and \hypergram{}~\cite{na2026hypergram}.
\textsuperscript{\S} published numbers (reference only);
\textsuperscript{$\star$} authors' released checkpoint, evaluated on our protocol;
\textsuperscript{$\dag$} trained from the authors' released code at their recipe.
}
\label{tab:other_video_results}
\footnotesize
\setlength{\tabcolsep}{3pt}
\renewcommand{\arraystretch}{1.05}
\resizebox{1.7\columnwidth}{!}{%
\begin{tabularx}{\textwidth}{
    @{}
    l
    c
    *{12}{>{\centering\arraybackslash}X}
    @{}
}
\toprule
\multirow{3}{*}{Method}
& \multirow{3}{*}{Modalities}
& \multicolumn{4}{c}{\textbf{DiDeMo}}
& \multicolumn{4}{c}{\textbf{ActivityNet}}
& \multicolumn{4}{c}{\textbf{VATEX}} \\
\cmidrule(lr){3-6}
\cmidrule(lr){7-10}
\cmidrule(lr){11-14}

& & \multicolumn{2}{c}{T2V}
& \multicolumn{2}{c}{V2T}
& \multicolumn{2}{c}{T2V}
& \multicolumn{2}{c}{V2T}
& \multicolumn{2}{c}{T2V}
& \multicolumn{2}{c}{V2T} \\
\cmidrule(lr){3-4}
\cmidrule(lr){5-6}
\cmidrule(lr){7-8}
\cmidrule(lr){9-10}
\cmidrule(lr){11-12}
\cmidrule(lr){13-14}

& & R@1 & R@10 & R@1 & R@10
& R@1 & R@10 & R@1 & R@10
& R@1 & R@10 & R@1 & R@10 \\
\midrule

UMT~\cite{Li2023UnmaskedTT}\textsuperscript{\S}
& T--V
& 34.0 & 68.7 & 34.0 & 68.7
& 31.9 & 72.0 & 31.9 & 72.0
& - & - & - & - \\

VideoCoCa~\cite{yan2022videococa}\textsuperscript{\S}
& T--V
& - & - & - & -
& 34.5 & 76.6 & 34.5 & 76.6
& 53.2 & - & - & - \\

HiTeA~\cite{Ye2022HiTeAHT}\textsuperscript{\S}
& T--V
& 43.2 & 79.0 & 56.5 & -
& - & - & - & -
& - & - & - & - \\

TVTSv2~\cite{Zeng2023TVTSv2LO}\textsuperscript{\S}
& T--V
& 34.6 & 71.5 & 34.6 & 71.5
& - & - & - & -
& - & - & - & - \\

UMT-L~\cite{Li2023UnmaskedTT}\textsuperscript{\S}
& T--V
& 48.6 & 79.0 & 24.9 & -
& 41.9 & - & 39.4 & -
& - & - & - & - \\

InternVideo-L~\cite{Wang2022InternVideoGV}\textsuperscript{\S}
& T--V
& 31.5 & - & 33.5 & -
& 30.7 & - & 31.4 & -
& 49.5 & - & 69.5 & - \\

OmniVL~\cite{Wang2022OmniVLOF}\textsuperscript{\S}
& T--V
& 40.6 & 74.3 & 33.3 & 68.5
& - & - & - & -
& - & - & - & - \\

ViCLIP~\cite{Wang2023InternVidAL}\textsuperscript{\S}
& T--V
& 18.4 & - & 27.9 & -
& 15.1 & - & 24.0 & -
& - & - & - & - \\

LanguageBind~\cite{Zhu2023LanguageBindEV}\textsuperscript{\S}
& T--V
& 39.9 & 74.6 & 39.8 & 76.2
& 41.0 & 80.0 & 39.1 & 81.1
& - & - & - & - \\

mPLUG-2~\cite{Xu2023mPLUG2AM}\textsuperscript{\S}
& T--V
& 45.7 & 71.1 & - & -
& - & - & - & -
& - & - & - & - \\

VideoPrism-b~\cite{Zhao2024VideoPrismAF}\textsuperscript{\S}
& T--V
& - & - & - & -
& 49.6 & - & 47.9 & -
& 62.5 & - & 76.2 & - \\

\addlinespace[2pt]

\pmrl{}~\cite{liu2026principled}\textsuperscript{$\star$}
& T--V
& 53.0 & 82.1 & 50.5 & 81.1
& 53.8 & 87.9 & 48.0 & 84.8
& 87.7 & 99.1 & 85.4 & 98.6 \\

\gram{}~\cite{cicchetti2025gramian}\textsuperscript{$\star$}
& T--V
& 51.3 & 78.3 & 49.0 & 77.5
& 55.5 & 88.2 & 49.3 & 84.1
& 87.2 & 99.1 & 84.7 & 98.1 \\

\rowcolor{gray!15}
-\quad w \hyvol
& T--V
& 53.8 \textbf{\textcolor{green!50!black}{(+2.5)}}
& 78.5 \textbf{\textcolor{green!50!black}{(+0.2)}}
& 53.8 \textbf{\textcolor{green!50!black}{(+4.8)}}
& 78.4 \textbf{\textcolor{green!50!black}{(+0.9)}}
& 56.2 \textbf{\textcolor{green!50!black}{(+0.7)}}
& 88.6 \textbf{\textcolor{green!50!black}{(+0.4)}}
& 56.2 \textbf{\textcolor{green!50!black}{(+6.9)}}
& 85.2 \textbf{\textcolor{green!50!black}{(+1.1)}}
& 92.1 \textbf{\textcolor{green!50!black}{(+4.9)}}
& 99.1 \textbf{\textcolor{green!50!black}{(+0.0)}}
& 92.3 \textbf{\textcolor{green!50!black}{(+7.6)}}
& 98.1 \textbf{\textcolor{green!50!black}{(+0.0)}} \\

\hypergram{}~\cite{na2026hypergram}\textsuperscript{$\dag$}
& T--V
& 50.0 & 77.5 & 50.7 & 78.1
& 56.1 & 88.4 & 50.5 & 87.9
& 88.4 & 99.1 & 86.3 & 97.6 \\

\rowcolor{gray!15}
\quad w \hyvol
& T--V
& 53.6 \textbf{\textcolor{green!50!black}{(+3.6)}}
& 79.2 \textbf{\textcolor{green!50!black}{(+1.7)}}
& 53.8 \textbf{\textcolor{green!50!black}{(+3.1)}}
& 80.4 \textbf{\textcolor{green!50!black}{(+2.3)}}
& 56.7 \textbf{\textcolor{green!50!black}{(+0.6)}}
& 88.7 \textbf{\textcolor{green!50!black}{(+0.3)}}
& 55.8 \textbf{\textcolor{green!50!black}{(+5.3)}}
& 88.5 \textbf{\textcolor{green!50!black}{(+0.6)}}
& 92.6 \textbf{\textcolor{green!50!black}{(+4.2)}}
& 99.1 \textbf{\textcolor{green!50!black}{(+0.0)}}
& 92.4 \textbf{\textcolor{green!50!black}{(+6.1)}}
& 99.3 \textbf{\textcolor{green!50!black}{(+1.7)}} \\

\midrule

\vast{}\cite{chen2023vast}\textsuperscript{\S}
& T--VA
& 49.5 & 76.9 & 48.2 & 78.6
& 51.4 & 83.6 & 46.8 & 77.4
& 80.0 & 95.7 & 77.1 & 95.2 \\

\trianglemodel{}~\cite{cicchetti2026triangle}\textsuperscript{$\star$}
& T--VA
& 51.0 & 75.5 & 49.4 & 79.2
& 56.0 & 83.8 & 52.3 & 87.2
& 88.9 & 99.8 & 86.5 & 99.1 \\

\pmrl{}~\cite{liu2026principled}\textsuperscript{$\star$}
& T--VA
& 52.5 & 79.9 & 49.3 & 80.4
& 54.1 & 85.3 & 48.6 & 84.1
& 88.9 & 99.8 & 86.1 & 99.1 \\

\gram{}~\cite{cicchetti2025gramian}\textsuperscript{$\star$}
& T--VA
& 50.6 & 77.7 & 49.0 & 76.8
& 56.4 & 87.7 & 49.7 & 84.3
& 88.4 & 99.1 & 86.5 & 98.4 \\

\rowcolor{gray!15}
-\quad w \hyvol
& T--VA
& 54.6 \textbf{\textcolor{green!50!black}{(+4.0)}}
& 77.9 \textbf{\textcolor{green!50!black}{(+0.2)}}
& 54.1 \textbf{\textcolor{green!50!black}{(+5.1)}}
& 78.8 \textbf{\textcolor{green!50!black}{(+2.0)}}
& 56.9 \textbf{\textcolor{green!50!black}{(+0.5)}}
& 87.9 \textbf{\textcolor{green!50!black}{(+0.2)}}
& 56.6 \textbf{\textcolor{green!50!black}{(+6.9)}}
& 84.8 \textbf{\textcolor{green!50!black}{(+0.5)}}
& 93.0 \textbf{\textcolor{green!50!black}{(+4.6)}}
& 99.3 \textbf{\textcolor{green!50!black}{(+0.2)}}
& 92.8 \textbf{\textcolor{green!50!black}{(+6.3)}}
& 98.4 \textbf{\textcolor{green!50!black}{(+0.0)}} \\

\hypergram{}~\cite{na2026hypergram}\textsuperscript{$\dag$}
& T--VA
& 50.4 & 77.8 & 50.8 & 78.8
& 56.8 & 88.9 & 50.9 & 88.3
& 88.6 & 99.5 & 88.4 & 98.6 \\

\rowcolor{gray!15}
\quad w \hyvol
& T--VA
& 54.2 \textbf{\textcolor{green!50!black}{(+3.8)}}
& 78.6 \textbf{\textcolor{green!50!black}{(+0.8)}}
& 54.7 \textbf{\textcolor{green!50!black}{(+3.9)}}
& 80.0 \textbf{\textcolor{green!50!black}{(+1.2)}}
& 57.8 \textbf{\textcolor{green!50!black}{(+1.0)}}
& 89.5 \textbf{\textcolor{green!50!black}{(+0.6)}}
& 56.4 \textbf{\textcolor{green!50!black}{(+5.5)}}
& 88.7 \textbf{\textcolor{green!50!black}{(+0.4)}}
& 93.0 \textbf{\textcolor{green!50!black}{(+4.4)}}
& 99.8 \textbf{\textcolor{green!50!black}{(+0.3)}}
& 92.9 \textbf{\textcolor{green!50!black}{(+4.5)}}
& 99.3 \textbf{\textcolor{green!50!black}{(+0.7)}} \\

\midrule

\vast{}~\cite{chen2023vast}\textsuperscript{\S}
& T--VAS
& - & - & - & -
& - & - & - & -
& 82.1 & 96.8 & 78.7 & 97.7 \\

\pmrl{}~\cite{liu2026principled}\textsuperscript{$\star$}
& T--VAS
& - & - & - & -
& - & - & - & -
& 89.6 & 98.8 & 87.0 & 98.4 \\

\gram{}~\cite{cicchetti2025gramian}\textsuperscript{$\star$}
& T--VAS
& - & - & - & -
& - & - & - & -
& 88.4 & 99.3 & 86.5 & 98.4 \\

\rowcolor{gray!15}
-\quad w \hyvol
& T--VAS
& - & - & - & -
& - & - & - & -
& 94.0 \textbf{\textcolor{green!50!black}{(+5.6)}}
& 100.0 \textbf{\textcolor{green!50!black}{(+0.7)}}
& 93.8 \textbf{\textcolor{green!50!black}{(+7.3)}}
& 98.8 \textbf{\textcolor{green!50!black}{(+0.4)}} \\

\hypergram{}~\cite{na2026hypergram}\textsuperscript{$\dag$}
& T--VAS
& - & - & - & -
& - & - & - & -
& 89.8 & 99.5 & 87.8 & 98.6 \\

\rowcolor{gray!15}
\quad w \hyvol
& T--VAS
& - & - & - & -
& - & - & - & -
& 93.3 \textbf{\textcolor{green!50!black}{(+3.5)}}
& 99.5 \textbf{\textcolor{green!50!black}{(+0.0)}}
& 93.9 \textbf{\textcolor{green!50!black}{(+6.1)}}
& 99.1 \textbf{\textcolor{green!50!black}{(+0.5)}} \\

\bottomrule
\end{tabularx}}
\end{table*}
\begin{table}[t]
\centering
\caption{
Zero-shot audio retrieval on AudioCaps~\cite{AUDIOCAPS} and audio--text classification on VGGSound-5K~\cite{VGGSOUND}.
\hyvol is applied on top of \gram{}~\cite{cicchetti2025gramian} and \hypergram{}~\cite{na2026hypergram}. \textsuperscript{\S} published numbers (reference only);
\textsuperscript{$\star$} authors' released checkpoint, evaluated on our protocol;
\textsuperscript{$\dag$} trained from the authors' released code at their recipe.
}
\label{tab:audio_results}
\scriptsize
\setlength{\tabcolsep}{3pt}
\renewcommand{\arraystretch}{1.05}

\begin{tabularx}{\columnwidth}{
    @{}
    l
    c
    *{4}{>{\centering\arraybackslash}X}
    @{}
}
\toprule
\multirow{3}{*}{Method}
& \multirow{3}{*}{Modalities}
& \multicolumn{2}{c}{\textbf{AudioCaps}}
& \multicolumn{2}{c}{\textbf{VGGSound-5K}} \\
\cmidrule(lr){3-4}
\cmidrule(lr){5-6}

& & \multicolumn{2}{c}{Retrieval}
& \multicolumn{2}{c}{Classification} \\
\cmidrule(lr){3-4}
\cmidrule(lr){5-6}

& & R@1 & R@10 & Acc@1 & Acc@10 \\
\midrule

\multicolumn{6}{l}{\emph{Audio-only and single-pair baselines}} \\

AVFIC~\cite{Nagrani2022LearningAM}\textsuperscript{\S}
& T--A
& 8.7 & 37.7
& - & - \\

ImageBind~\cite{Girdhar2023ImageBindOE}\textsuperscript{\S}
& T--A
& 9.3 & 42.3
& - & - \\

LanguageBind~\cite{Zhu2023LanguageBindEV}\textsuperscript{\S}
& T--A
& 19.7 & 67.6
& 23.8 & 57.1 \\

VIP-ANT~\cite{Zhao2021ConnectingTD}\textsuperscript{\S}
& T--A
& 27.7 & 37.7
& - & - \\

\vast{}\cite{chen2023vast}\textsuperscript{\S}
& T--A
& - & -
& 25.6 & 56.2 \\

LanguageBind~\cite{Zhu2023LanguageBindEV}\textsuperscript{\S}
& T--V
& - & -
& 37.2 & 62.0 \\

\vast{}\cite{chen2023vast}\textsuperscript{\S}
& T--V
& - & -
& 38.7 & 72.8 \\

\midrule

\multicolumn{6}{l}{\emph{Multimodal (audio $+$ video) methods}} \\

AVFIC~\cite{Nagrani2022LearningAM}\textsuperscript{\S}
& T--AV
& 10.6 & 45.2
& - & - \\

\vast{}\cite{chen2023vast}\textsuperscript{\S}
& T--AV
& 32.1 & 65.4
& 39.6 & 74.5 \\

\addlinespace[2pt]

\trianglemodel{}~\cite{cicchetti2026triangle}\textsuperscript{$\star$}
& T--AV
& 37.4 & 71.9
& 45.1 & 81.5 \\

\pmrl{}~\cite{liu2026principled}\textsuperscript{$\star$}
& T--AV
& 34.4 & 76.3
& 42.8 & 80.0 \\

\gram{}~\cite{cicchetti2025gramian}\textsuperscript{$\star$}
& T--AV
& 33.4 & 74.4
& 41.0 & 77.8 \\

\rowcolor{gray!15}
-\quad w \hyvol
& T--AV
& 35.7 \textbf{\textcolor{green!50!black}{(+2.3)}}
& 74.9 \textbf{\textcolor{green!50!black}{(+0.5)}}
& 40.9 \textbf{\textcolor{red!50!black}{(-0.1)}}
& 77.7 \textbf{\textcolor{red!50!black}{(-0.1)}} \\

\hypergram{}~\cite{na2026hypergram}\textsuperscript{$\dag$}
& T--AV
& 32.4 & 72.9
& 40.7 & 76.8 \\

\rowcolor{gray!15}
\quad w \hyvol
& T--AV
& 36.1 \textbf{\textcolor{green!50!black}{(+3.7)}}
& 76.8 \textbf{\textcolor{green!50!black}{(+3.9)}}
& 41.7 \textbf{\textcolor{green!50!black}{(+1.0)}}
& 76.6 \textbf{\textcolor{red!50!black}{(-0.2)}} \\

\bottomrule
\end{tabularx}
\end{table}
\noindent\textbf{Backbones and Pretraining.} All variants employ \vast{} modality encoders~\cite{chen2023vast}: EVA-CLIP ViT-g/14 for video~\cite{Sun2023EVACLIPIT}, BEATs for audio~\cite{beats2023}, and BERT-base for captions and subtitles~\cite{devlin2019bert}. We remove the VAST fusion layers, project each stream to $D=512$, and apply $\ell_2$ normalization. Models are trained for one epoch on a 150K-sample subset of VAST-27M, using two video frames, one 10-second audio clip, and a global batch size of 256. AdamW~\cite{loshchilov2019adamw} is used with a weight decay of $0.01$, learning rates of $2\times10^{-5}$ for the backbone and $5\times10^{-4}$ for the hypergraph parameters, two refinement layers, and $k=12$ as described in Sec.~\ref{sec:neighborhood}. The hypergraph is constructed independently within each mini-batch, so the batch size also defines the semantic-neighbor pool. The remaining hyperparameters are \(w_{\mathrm{doc}}=1.0\), \(w_{\mathrm{reg}}=0.1\), \(w_{\mathrm{itm}}=0.1\), semantic-edge dropout \(0.3\), and \(\epsilon=10^{-4}\) in Eq.~\eqref{eq:vicreg}; the temperature \(\tau\) is initialised from the backbone checkpoint and continues to be optimised. We perform aggregation in float32 for numerical stability. Each variant \hyvol{} adopts its published training recipe, identical data, encoders, optimization, sampling, and schedule, so that the margin isolates the module.

\noindent\textbf{Benchmarks and Evaluation Protocols.}
We evaluate zero-shot performance on six benchmarks: text--video retrieval on MSR-VTT~\cite{MSRVTT} (1,000 test clips), DiDeMo~\cite{DIDEMO} (1,003), ActivityNet~\cite{ACTIVITYNET} (4,917), and VATEX~\cite{wang2019vatex} (431); text--audio retrieval on AudioCaps~\cite{AUDIOCAPS} (700); and audio-visual classification on VGGSound-5K~\cite{VGGSOUND}. At inference, we sample eight video frames and one 10-second audio segment per clip; VGGSound uses embeddings of its 309 class labels. The $\mathrm{T}$--$\mathrm{V}$, $\mathrm{T}$--$\mathrm{VA}$, and $\mathrm{T}$--$\mathrm{VAS}$ settings progressively add video, audio, and subtitles to text. We evaluate all three on MSR-VTT and VATEX, and the first two on DiDeMo and ActivityNet. We report R@1 and R@10 for text-to-video (T2V), video-to-text (V2T), and text-to-audio retrieval, and top-1 and top-10 accuracy for VGGSound.

We compare against \trianglemodel{}~\cite{cicchetti2026triangle}, \pmrl{}~\cite{liu2026principled}, \gram{}~\cite{cicchetti2025gramian}, and \hypergram{}~\cite{na2026hypergram}. Each computes a joint geometric retrieval score solely from an individual sample's modality embeddings (Sec.~\ref{sec:method}), whereas \hyvol{} relaxes this restriction. All four are evaluated under the common protocol described in Sec.~\ref{sec:experiments}, enabling a fair comparison. At inference, the hypergraph is removed, and retrieval uses the original encoders and unmodified score $\nu_b$, with no message passing or interaction among test samples. The deployed architecture and scoring procedure therefore remain identical to the backbone. Additional computation is limited to training and mainly comprises a $B\times B$ caption-similarity matrix, hyperedge construction, and two shallow message-passing layers; wall-clock overhead is not measured.

\subsection{Main Results}
\label{sec:main_results}
\noindent\textbf{MSR-VTT.} Table~\ref{tab:msrvtt_results} presents positive R@1 changes for both backbones across all evaluated modality sets and retrieval directions. When using $\mathrm{T}$--$\mathrm{VAS}$, the addition of \hyvol{} to \gram{} increases T2V R@1 from $51.8$ to $55.8$ ($+4.0$) and V2T R@1 from $50.1$ to $58.1$ ($+8.0$). For \hypergram{}, the respective changes are $52.3\rightarrow56.1$ ($+3.8$) and $50.9\rightarrow59.2$ ($+8.3$). When only text and video are used, the T2V R@1 improvements are smaller: $51.9\rightarrow52.6$ for \gram{} and $50.8\rightarrow52.5$ for \hypergram{}. The corresponding V2T improvements are $+5.8$ and $+5.5$ points. On MSR-VTT, T2V R@1 improvement increases as audio and subtitles are incorporated for both backbones. 
% This trend aligns with the intended application of multi-stream document hyperedges. 
% However, this is a dataset-specific observation, as modality cardinality is not varied independently and the contribution of each edge type is not isolated in this experiment.

\noindent\textbf{DiDeMo, ActivityNet, and VATEX.} Table~\ref{tab:other_video_results} reports positive R@1 changes for all paired comparisons across the three datasets, though the magnitude of improvement varies. For DiDeMo, improvements range from $+2.5$ to $+5.1$ points across different backbones, modality sets, and retrieval directions. In ActivityNet, changes span from $+0.5$ to $+6.9$, with the largest increases observed in V2T for \gram{}. VATEX also exhibits positive changes, despite relatively high baseline backbone scores. Using $\mathrm{T}$--$\mathrm{VAS}$, \gram{} improves from $88.4$ to $94.0$ in T2V and from $86.5$ to $93.8$ in V2T. The corresponding \hypergram{} results increase from $89.8$ to $93.3$ and from $87.8$ to $93.9$. These findings confirm the direction of the measured changes under the paired training protocol. However, the results do not independently determine whether document hyperedges, semantic hyperedges, or their interaction is responsible for the observed gains.

\noindent\textbf{Audio retrieval and classification.}
On AudioCaps (Table~\ref{tab:audio_results}), \hyvol{} increases \gram{} R@1/R@10 from $33.4/74.4$ to $35.7/74.9$ and \hypergram{} from $32.4/72.9$ to $36.1/76.8$. In contrast, VGGSound-5K exhibits smaller and mixed changes: \gram{} top-1/top-10 accuracy shifts from $41.0/77.8$ to $40.9/77.7$, while \hypergram{} changes from $40.7/76.8$ to $41.7/76.6$. These four differences range from $-0.2$ to $+1.0$ points. Since each configuration is trained once and seed variance is not measured, we read the classification results as approximately unchanged. A plausible cause is the mismatch between caption-supervised neighbourhood construction and evaluation against short class labels. However, the current experiments do not isolate this factor, so this interpretation remains tentative.

\noindent\textbf{Cross-benchmark observations.} Across the video benchmarks, V2T R@1 gain exceeds T2V gain in 19 of 20 paired backbone–modality comparisons. The only exception is the T–V \hypergram{} result on DiDeMo, where the changes are +3.6 in T2V and +3.1 in V2T. This directional pattern aligns with the observation that \hyvol{} refines candidate-side representations while leaving caption embeddings outside the hypergraph. 

% Nevertheless, as the experiment does not include a direction-specific intervention, this pattern should be interpreted as descriptive rather than causal.

The effect of modality cardinality varies across datasets: unlike
MSR-VTT, VATEX shows no monotonic trend across
\(\mathrm{T}\)--\(\mathrm{V}\), \(\mathrm{T}\)--\(\mathrm{VA}\), and
\(\mathrm{T}\)--\(\mathrm{VAS}\). Additionally, \hypergram{} demonstrates greater improvements than \gram{} in several scenarios, though this pattern is not consistent on VATEX. These findings indicate that the effectiveness of the refinement depends on both the backbone and the benchmark. Our experiments vary two factors: the neighbourhood size \(k\) (Sec.~\ref{sec:neighborhood}) and the fraction of gallery clips with a missing stream (Sec.~\ref{sec:missing}). 
% They do not vary the components
% of the module itself, so these differences cannot be attributed here to a particular edge type or to the scoring geometry; we record this among the limitations in Sec.~\ref{sec:limitations}.
\begin{figure*}[t]
    \centering
    \includegraphics[width=0.693\textwidth]{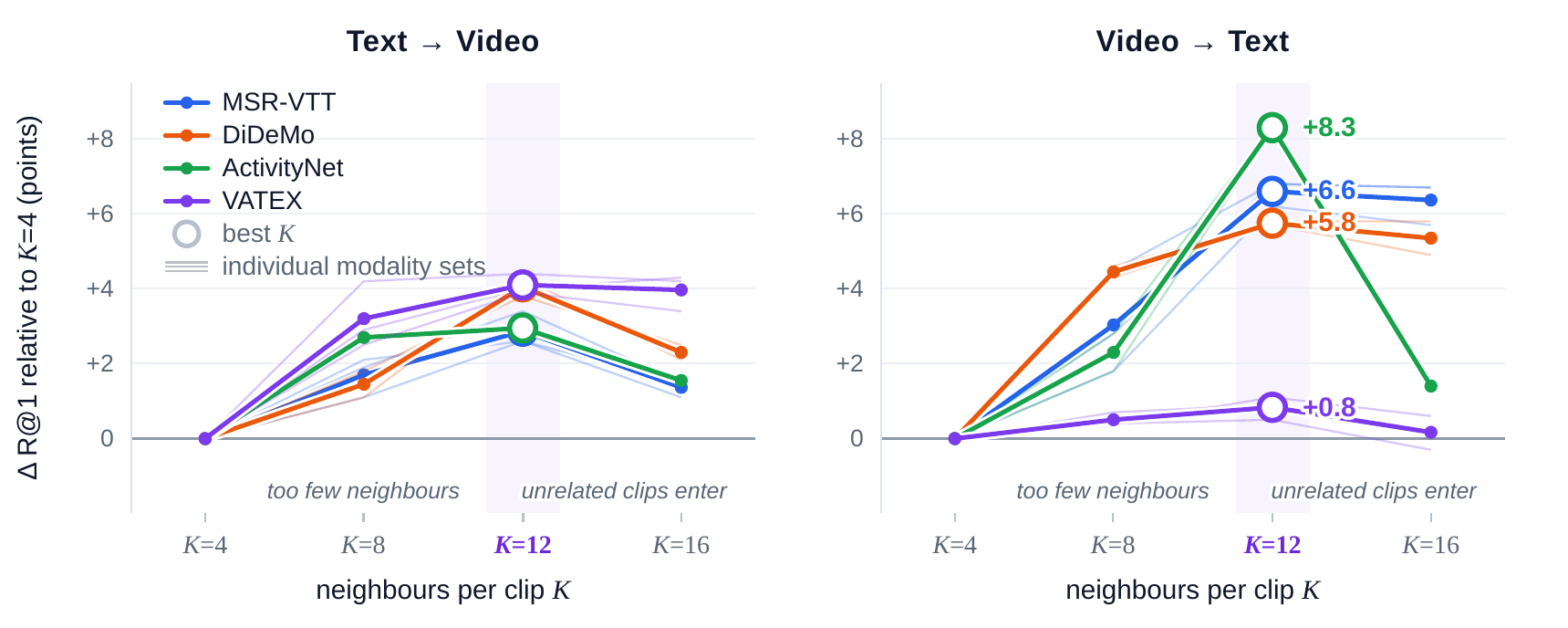}
    \caption{Sensitivity to the number of semantic neighbours $k$. Values are
    R@1 changes relative to $k=4$. Dark curves aggregate each benchmark and
    faint curves show individual modality sets; outlined markers denote the
    best tested $k$.  The reported configuration uses \hyvol{} on top of \gram{} backbone.}
    \label{fig:k_ablation}
\end{figure*}
\begin{table}[t]
\centering
\caption{Retrieval under missing modalities (R@1). A fraction $r$ of gallery
clips loses one randomly chosen modality (deterministic per-clip masks,
identical across methods; $r{=}0$ reproduces the main protocol). \hyvol{} is
applied on top of \gram{}~\cite{cicchetti2025gramian}. MSR-VTT and VATEX use the
T--VAS modality set; DiDeMo and ActivityNet use T--VA. ANet: ActivityNet.
\textsuperscript{$\star$}~authors' released checkpoint, evaluated on our protocol. }
\label{tab:missing_modality}
\setlength{\tabcolsep}{2pt}
\renewcommand{\arraystretch}{1.0}
\resizebox{\columnwidth}{!}{%
\begin{tabular}{@{}l l ccccc c ccccc@{}}
\toprule
 & & \multicolumn{5}{c}{T$\to$V} & & \multicolumn{5}{c}{V$\to$T} \\
\cmidrule(lr){3-7} \cmidrule(lr){9-13}
 & Method & 0\% & 25\% & 50\% & 75\% & 90\% & & 0\% & 25\% & 50\% & 75\% & 90\% \\
\midrule
\multirow{3}{*}{MSR-VTT}
 & \gram{}\textsuperscript{$\star$} & 51.8 & 49.2 & 44.0 & 41.4 & 39.5 & & 50.1 & 46.5 & 42.4 & 38.1 & 36.4 \\
 & - w \hyvol{} & \textbf{55.8} & \textbf{52.7} & \textbf{47.2} & \textbf{41.6} & 39.3 & & \textbf{58.1} & \textbf{54.1} & \textbf{48.0} & \textbf{42.6} & \textbf{39.4} \\
 & \textcolor{green!50!black}{\textit{margin}} & \textcolor{green!50!black}{$+4.0$} & \textcolor{green!50!black}{$+3.5$} & \textcolor{green!50!black}{$+3.2$} & \textcolor{green!50!black}{$+0.2$} & \textcolor{red!60!black}{$-0.2$} & & \textcolor{green!50!black}{$+8.0$} & \textcolor{green!50!black}{$+7.6$} & \textcolor{green!50!black}{$+5.6$} & \textcolor{green!50!black}{$+4.5$} & \textcolor{green!50!black}{$+3.0$} \\
\cmidrule(l){2-13}
\multirow{3}{*}{DiDeMo}
 & \gram{}\textsuperscript{$\star$} & 50.6 & 45.5 & 40.6 & 36.6 & 33.5 & & 49.0 & 42.5 & 36.7 & 32.1 & 28.6 \\
 & - w \hyvol{} & \textbf{54.6} & \textbf{47.5} & \textbf{42.4} & \textbf{38.1} & \textbf{35.2} & & \textbf{54.1} & \textbf{47.5} & \textbf{41.1} & \textbf{36.1} & \textbf{31.2} \\
 & \textcolor{green!50!black}{\textit{margin}} & \textcolor{green!50!black}{$+4.0$} & \textcolor{green!50!black}{$+2.0$} & \textcolor{green!50!black}{$+1.8$} & \textcolor{green!50!black}{$+1.5$} & \textcolor{green!50!black}{$+1.7$} & & \textcolor{green!50!black}{$+5.1$} & \textcolor{green!50!black}{$+5.0$} & \textcolor{green!50!black}{$+4.4$} & \textcolor{green!50!black}{$+4.0$} & \textcolor{green!50!black}{$+2.6$} \\
\cmidrule(l){2-13}
\multirow{3}{*}{ANet}
 & \gram{}\textsuperscript{$\star$} & 56.4 & 50.4 & 44.4 & 38.1 & 34.2 & & 49.7 & 43.9 & 38.4 & 31.7 & 28.2 \\
 & - w \hyvol{} & \textbf{56.9} & 49.0 & 43.5 & 37.8 & 34.2 & & \textbf{56.6} & \textbf{48.5} & \textbf{41.6} & \textbf{32.9} & \textbf{29.0} \\
 & \textcolor{green!50!black}{\textit{margin}} & \textcolor{green!50!black}{$+0.5$} & \textcolor{red!60!black}{$-1.4$} & \textcolor{red!60!black}{$-0.9$} & \textcolor{red!60!black}{$-0.3$} & $\pm0.0$ & & \textcolor{green!50!black}{$+6.9$} & \textcolor{green!50!black}{$+4.6$} & \textcolor{green!50!black}{$+3.2$} & \textcolor{green!50!black}{$+1.2$} & \textcolor{green!50!black}{$+0.8$} \\
\cmidrule(l){2-13}
\multirow{3}{*}{VATEX}
 & \gram{}\textsuperscript{$\star$} & 88.4 & 80.7 & 74.5 & 67.7 & 64.5 & & 86.5 & 79.1 & 73.5 & 67.3 & 64.0 \\
 & - w \hyvol{} & \textbf{94.0} & \textbf{85.8} & \textbf{78.0} & \textbf{70.8} & \textbf{66.4} & & \textbf{93.8} & \textbf{84.9} & \textbf{77.3} & \textbf{69.8} & \textbf{65.2} \\
 & \textcolor{green!50!black}{\textit{margin}} & \textcolor{green!50!black}{$+5.6$} & \textcolor{green!50!black}{$+5.1$} & \textcolor{green!50!black}{$+3.5$} & \textcolor{green!50!black}{$+3.1$} & \textcolor{green!50!black}{$+1.9$} & & \textcolor{green!50!black}{$+7.3$} & \textcolor{green!50!black}{$+5.8$} & \textcolor{green!50!black}{$+3.8$} & \textcolor{green!50!black}{$+2.5$} & \textcolor{green!50!black}{$+1.2$} \\
\bottomrule
\end{tabular}%
}
\end{table}
\subsubsection{Neighbourhood-Size Sensitivity}
\label{sec:neighborhood}
Figure~\ref{fig:k_ablation} presents a comparison of $k\in\{4,8,12,16\}$. Performance improves as $k$ increases from 4 to 12, with $k=12$ yielding the highest benchmark averages for both retrieval directions. Compared to $k=4$, V2T R@1 at $k=12$ increases by $6.6$ on MSR-VTT, $5.8$ on DiDeMo, $8.3$ on ActivityNet, and $0.8$ on VATEX. Increasing $k$ to $16$ does not provide further average improvement and substantially reduces ActivityNet V2T performance. Consequently, $k=12$ is selected for the main experiments. This pattern aligns with a neighborhood-density trade-off: small $k$ results in few cross-document edges under mutual-neighbor filtering, while large $k$ may include less similar captions. However, this interpretation is tentative, as the number, similarity, and semantic precision of the retained edges are not directly measured. Under the paired training protocol, \hyvol{} increases R@1 across all five retrieval benchmarks, although the magnitude of improvement varies by dataset, retrieval direction, modality set, and backbone. The changes observed on VGGSound-5K are minimal. 
% Since the hypergraph is removed after training, the deployed retrieval pathway remains unchanged.

\subsubsection{Missing-Modality Ablation}
\label{sec:missing}
Table~\ref{tab:missing_modality} presents results where one randomly selected modality is removed from a fraction \(r\) of gallery clips, applying identical per-clip masks across both methods. When \(r=0\), the main protocol is recovered. Missing streams are not imputed, and Eq.~\eqref{eq:masked_gram} evaluates each document based solely on its observed modalities. At \(r=0.9\), \gram{} exhibits a decrease of \(13.7\) V2T R@1 points on MSR-VTT and \(22.5\) points on VATEX compared to \(r=0\). The impact of \hyvol{} differs depending on the retrieval direction. Its V2T margin remains positive across all 20 settings (ranging from \(+0.8\) to \(+8.0\)), while its T2V margin is slightly negative in four cases: MSR-VTT at
\(r=0.9\) and ActivityNet at \(r=0.25\) to \(0.75\). This asymmetry aligns with \hyvol{} refining only document-side embeddings (see Sec.~\ref{sec:traintest}), resulting in greater benefits when the document is used as the query. Masking removes unavailable modalities from document hyperedges but retains caption-derived semantic edges. Across all four benchmarks, the V2T margin decreases monotonically with \(r\) but remains positive: \(+8.0\rightarrow+3.0\) on MSR-VTT, \(+6.9\rightarrow+0.8\) on ActivityNet, \(+7.3\rightarrow+1.2\) on VATEX, and \(+5.1\rightarrow+2.6\) on DiDeMo. This trend aligns with the hypothesis that masking reduces the arity of document hyperedges while leaving caption-derived semantic hyperedges unchanged. Nevertheless, since the remaining document hyperedges remain active, the residual margin at \(r=0.9\) cannot be attributed exclusively to cross-document refinement in the absence of an edge-type ablation. 
% This trend may suggest a greater reliance on within-document fusion in the latter datasets; however, separating the two edge types is necessary to directly test this hypothesis.

\section{Conclusion}
We introduced \hyvol{}, a training-time hypergraph regularizer that integrates within-document and cross-document semantic relations to refine modality embeddings prior to Gramian-volume alignment. \hyvol{} is compatible with both \gram{} and \hypergram{}, accommodates missing streams without imputation, and incurs no additional inference-time module or retrieval cost. Evaluated on six zero-shot benchmarks, \hyvol{} increases R@1 across all five retrieval tasks, achieving gains of up to \(+8.3\), while VGGSound-5K classification performance remains stable. Under missing-modality masking, the V2T margin remains positive in all 20 settings, although four T2V cases exhibit slight negative margins. These findings indicate that batch-level candidate relations can enhance candidate-local geometric alignment while maintaining the efficiency of the deployed retrieval pathway.

 %\ \input{sec/1_intro}
 %\ \input{sec/2_formatting}
 %\ \input{sec/3_method}
{
    \small
    \bibliographystyle{ieeenat_fullname}
    \bibliography{main}
}

\clearpage
\appendix
\setcounter{table}{0}\renewcommand{\thetable}{S\arabic{table}}
\setcounter{figure}{0}\renewcommand{\thefigure}{S\arabic{figure}}
\setcounter{equation}{0}\renewcommand{\theequation}{S\arabic{equation}}
\section{Use of Large Language Models}
\label{sec:supp-llm}
% -----------------------------------------------------------------------------
We used large language models solely as writing aids---to polish grammar,
improve clarity and word choice, and tighten length. All substantive
contributions, including the research idea, method design, experimental
protocol, analyses, and the numerical results, are the authors' own. Every
model-assisted edit was verified by the authors against our experimental
records, and the authors take full responsibility for the entire paper and its
supplementary material.

% -----------------------------------------------------------------------------
\section{Reproducibility}
\label{sec:supp-repro}
% -----------------------------------------------------------------------------
We will release the training and evaluation code, configuration files, the
scripts that generate our tables and figures, and the trained \hyvol{} weights
for the reported configuration. All \hyvol{} results use a single configuration
across all six benchmarks---five retrieval tasks and one audio--visual
classification task---with no benchmark-specific learning rate, temperature,
neighbourhood size, or schedule.

We evaluate \gram{},
\trianglemodel{}, and \pmrl{} using the checkpoints released by their
respective authors. As \hypergram{} does not provide a checkpoint, we train it from the authors' released code following their
published recipe. All evaluated methods use the \vast{} foundation model as the initial backbone and are assessed within the same evaluation environment and retrieval protocol. Published numbers are retained solely as reference values in the main paper.

\section{Implementation Details}
\label{sec:supp-implementation}
% -----------------------------------------------------------------------------

Table~\ref{tab:hparams} lists the reported \hyvol{} hyperparameters. The same
configuration is used for all six benchmarks---five retrieval tasks and one
audio--visual classification task---with no benchmark-specific tuning. The
hypergraph parameters are optimised with a
larger learning rate ($5\times10^{-4}$) than the encoders ($2\times10^{-5}$),
and all refinement operations run in \texttt{float32} for numerical stability
even when the encoders run in fp16. The semantic hyperedges are constructed
per mini-batch on the per-GPU shard ($B_{\mathrm{shard}}=64$ of the global
$B=256$),

\begin{table}[t]
\centering
\footnotesize
\renewcommand{\arraystretch}{1.15}
\caption{\hyvol hyperparameters. The same configuration is used for all six
benchmarks (five retrieval, one classification). $B_{\mathrm{shard}}=64$ is the
per-GPU batch of the global $B=256$.}
\label{tab:hparams}
\begin{tabular*}{\columnwidth}{@{\extracolsep{\fill}}llr@{}}
\toprule
Group & Hyperparameter & Value \\
\midrule
\multirow{5}{*}{Hypergraph}
 & neighbours $k$ (\texttt{knn\_k})              & $12$ \\
 & adaptive cap $\lfloor B_{\mathrm{shard}}/4\rfloor$ & $16$ \\
 & edge dropout $p_{\mathrm{drop}}$              & $0.3$ \\
 & similarity floor $\sigma_{\mathrm{std}}$      & disabled \\
 & semantic edges                                & train only \\
\midrule
\multirow{3}{*}{Edge attention}
 & feature dim $K$                               & $512$ \\
 & LeakyReLU slope                               & $0.2$ \\
 & $\mathbf{a}$ init                             & $\mathcal{N}(0,0.1^2)$ \\
\midrule
\multirow{3}{*}{GatedHGNN}
 & layers $L$                                    & $2$ \\
 & gate init $g_l$                               & $1.0$ \\
 & activation                                    & GELU \\
\midrule
\multirow{4}{*}{Loss}
 & temperature $\tau$ (learned)                  & $0.07$ \\
 & label smoothing                               & $0.1$ \\
 & doc weight $w_{\mathrm{doc}}$                 & $1.0$ \\
 & variance weight $w_{\mathrm{reg}}$            & $0.1$ \\
\midrule
\multirow{7}{*}{Training}
 & encoder lr $\eta$                             & $2\times10^{-5}$ \\
 & graph lr $\eta_{\mathrm{graph}}$              & $5\times10^{-4}$ \\
 & global batch $B$                              & $256$ \\
 & \quad per GPU ($\times4$)                     & $64$ \\
 & epochs / frames                               & $1$ / $2$ \\
 & audio per clip                                & one $10$\,s clip \\
 & precision                                     & fp16 (graph fp32) \\
\bottomrule
\end{tabular*}
\end{table}

\section{Dataset Statistics}
\label{sec:supp-data}
% -----------------------------------------------------------------------------
\begin{table}[t]
\centering
\caption{Dataset statistics for the evaluated protocol. $V$, $A$, and $S$
denote video, audio, and subtitles. Eight frames and one 10\,s audio segment
are sampled per clip. VGGSound-5K is a 309-way audio--visual classification
benchmark; the remaining datasets are used for retrieval. VATEX is evaluated
on the 431 test clips that remain downloadable, so its absolute recalls are
compared only within the common gallery used in this work.}
\label{tab:supp-datastats}
\scriptsize
\setlength{\tabcolsep}{4pt}
\begin{tabular}{@{}lcccccc@{}}
\toprule
Dataset & Streams & Train & Val & Test & Frames & Task \\
\midrule
MSR-VTT\textsuperscript{$\dag$} & $V,A,S$ & 9{,}000  & --      & 1{,}000 & 8 & Retrieval \\
DiDeMo           & $V,A$   & 8{,}394  & 1{,}065 & 1{,}003 & 8 & Retrieval \\
ActivityNet & $V,A$   & 10{,}009 & --      & 4{,}917 & 8 & Retrieval \\
VATEX     & $V,A,S$ & 14{,}060 & --      & \phantom{0,}431 & 8 & Retrieval \\
AudioCaps     & $V,A$   & --       & --      & \phantom{0,}700 & 8 & Retrieval \\
\midrule
VGGSound-5K    & $V,A$   & --       & --      & 5{,}000 & 8 & Classification \\
\bottomrule
\end{tabular}
\end{table}

\paragraph{Continued-pretraining corpus.}
Our \hyvol{} runs, controls, ablations, and \hypergram{} retraining use the same nominal 150K-clip subset of \vast{}-27M. At the time of our experiments, $136{,}674$ of
these clips remained downloadable and formed the effective corpus for the runs
we performed. Training uses one epoch, and the resulting logs contain 519
optimizer steps. The released \gram{},
\trianglemodel{}, and \pmrl{} checkpoints are evaluated as provided, so the
exact set of clips available when those checkpoints were originally trained
cannot be reconstructed from the artifacts alone.

\paragraph{Modality availability.}
The retrieval benchmarks cover two observed arities: MSR-VTT and
VATEX use video, audio, and subtitles, whereas DiDeMo, ActivityNet, and
AudioCaps use video and audio. VGGSound-5K, the
audio--visual classification benchmark, likewise uses video and audio, with its
309 class labels serving as text queries.

% -----------------------------------------------------------------------------
\section{Additional Results and Analysis}
\label{sec:supp-additional}
% -----------------------------------------------------------------------------

\subsection{Evaluation-Environment Audit}
\label{sec:supp-environment}

Table~\ref{tab:supp-gram-audit} scores the released
\gram{} checkpoint in two environments---its
published setting and ours---together with our \gram{} reproduction trained from
the released code and scored in our environment. Moving the \emph{same}
checkpoint from its published environment to ours changes text-to-video R@1 by
up to $3.0$ points on MSR-VTT ($54.8 \to 51.8$ on T--VAS), whereas re-scoring a
fixed checkpoint within a single environment is deterministic and leaves the
metric unchanged. This gap motivates the main paper's use of artifacts scored
under one environment, with published values shown only for reference.

The offset is not a constant that could simply be added back. Across MSR-VTT,
DiDeMo, and ActivityNet the released checkpoint's text-to-video R@1 falls
$0.9$ to $3.6$ points \emph{below} its published numbers under our protocol,
and our reproduction trained from the released code likewise falls below
published, by up to $5.3$ R@1. On VATEX the direction reverses: on the same $431$-clip gallery, both the
released checkpoint and our reproduction score \emph{above} the published
numbers---by up to $6.1$ and $6.8$ R@1, respectively---under our protocol. Because the sign and magnitude
of the discrepancy vary by benchmark, no single additive correction reconciles
the two environments, which is why we compare only artifacts scored under one
environment rather than adjusting published values.

\begin{table*}[t]
\centering
\caption{Evaluation-environment audit for \gram{} across all benchmarks (R@1/R@10, both directions). \emph{Weights}: the released
checkpoint versus our reproduction trained from the released code
(\emph{retrained}); \emph{Environment}: the paper's published numbers versus our
protocol. Differences between the published and our-protocol columns reflect the
evaluation environment, motivating a single-environment comparison.}
\label{tab:supp-gram-audit}
\scriptsize
\setlength{\tabcolsep}{2.2pt}
\renewcommand{\arraystretch}{1.05}
\begin{tabular}{@{}ll cc cc cc c cc cc cc@{}}
\toprule
& & \multicolumn{6}{c}{Text $\to$ Video} & & \multicolumn{6}{c}{Video $\to$ Text} \\
\cmidrule(lr){3-8}\cmidrule(lr){10-15}
\emph{Weights}
& & \multicolumn{2}{c}{released} & \multicolumn{2}{c}{released} & \multicolumn{2}{c}{retrained}
& & \multicolumn{2}{c}{released} & \multicolumn{2}{c}{released} & \multicolumn{2}{c}{retrained} \\
\emph{Environment}
& & \multicolumn{2}{c}{published} & \multicolumn{2}{c}{ours} & \multicolumn{2}{c}{ours}
& & \multicolumn{2}{c}{published} & \multicolumn{2}{c}{ours} & \multicolumn{2}{c}{ours} \\
\cmidrule(lr){3-4}\cmidrule(lr){5-6}\cmidrule(lr){7-8}\cmidrule(lr){10-11}\cmidrule(lr){12-13}\cmidrule(lr){14-15}
Dataset & Mode & R@1 & R@10 & R@1 & R@10 & R@1 & R@10 & & R@1 & R@10 & R@1 & R@10 & R@1 & R@10 \\
\midrule
\multirow{3}{*}{MSR-VTT}
 & T--V   & 52.8 & 82.9 & 51.9 & 82.4 & 50.6 & 81.8 & & 49.5 & 81.7 & 48.7 & 80.7 & 46.7 & 79.0 \\
 & T--VA  & 54.2 & 83.9 & 52.0 & 82.0 & 51.5 & 81.5 & & 50.5 & 82.2 & 49.3 & 80.7 & 47.1 & 79.6 \\
 & T--VAS & 54.8 & 82.9 & 51.8 & 81.9 & 52.2 & 82.6 & & 52.9 & 82.9 & 50.1 & 80.3 & 49.0 & 79.7 \\
\midrule
\multirow{2}{*}{DiDeMo}
 & T--V   & 54.0 & 80.7 & 51.3 & 78.3 & 49.8 & 77.4 & & 52.3 & 80.3 & 49.0 & 77.5 & 49.8 & 77.2 \\
 & T--VA  & 54.2 & 79.3 & 50.6 & 77.7 & 50.6 & 77.0 & & 52.2 & 78.9 & 49.0 & 76.8 & 49.5 & 77.4 \\
\midrule
\multirow{2}{*}{ActivityNet}
 & T--V   & 58.9 & 91.2 & 55.5 & 88.2 & 53.6 & 87.0 & & 50.9 & 85.4 & 49.3 & 84.1 & 48.6 & 81.7 \\
 & T--VA  & 59.0 & 91.1 & 56.4 & 87.7 & 54.6 & 85.9 & & 50.4 & 85.8 & 49.7 & 84.3 & 48.5 & 81.7 \\
\midrule
\multirow{3}{*}{VATEX}
 & T--V   & 81.1 & 99.5 & 87.2 & 99.1 & 87.9 & 99.3 & & 79.0 & 98.3 & 84.7 & 98.1 & 84.9 & 98.6 \\
 & T--VA  & 83.9 & 98.6 & 88.4 & 99.1 & 89.8 & 99.8 & & 79.2 & 99.0 & 86.5 & 98.4 & 86.8 & 98.6 \\
 & T--VAS & 83.5 & 98.8 & 88.4 & 99.3 & 88.6 & 99.3 & & 82.7 & 98.1 & 86.5 & 98.4 & 86.1 & 98.4 \\
\bottomrule
\end{tabular}
\end{table*}

For HyperGRAM the audit compares the published numbers with our reproduction
trained from the released code and scored in our environment. The reproduction
falls below the published text-to-video R@1, and the gap widens with the
modality set on MSR-VTT---from $2.4$ points on T--V to $4.3$ on T--VAS---while
staying under $1.5$ points on DiDeMo and ActivityNet. On VATEX the direction
reverses: evaluated on the same $431$-clip gallery, our reproduction scores
$9.9$ R@1 \emph{above} the published number. Because the sign and size of the
gap vary by benchmark, we rely on artifacts scored under a single environment
rather than on cross-source published values.

Table~\ref{tab:supp-triangle-audit} repeats the audit for \trianglemodel{},
whose area-based aggregator we evaluate on the four video--text benchmarks in the
T--VA setting; the pattern matches the volume backbones. Under our protocol the
released \trianglemodel{} checkpoint's text-to-video R@1 falls $3.7$ to $4.7$
points below the published numbers on MSR-VTT, DiDeMo, and ActivityNet, and our
reproduction trained from the released code falls further, by up to $8.4$ R@1. On
VATEX the direction reverses on the same $431$-clip gallery, where both the
released checkpoint and our reproduction score $5.0$ R@1 above the published
number. The \trianglemodel{} paper reports R@1 only, so the published R@10 is
omitted. As for \gram{} and HyperGRAM, the sign and magnitude of the gap vary by
benchmark, so published values from different sources cannot be compared
directly---motivating the single-environment protocol used in the main paper. 

\begin{table*}[t]
\centering
\caption{Evaluation-environment audit for \trianglemodel{} on the four
video--text benchmarks (T--VA, R@1/R@10, both directions).
\emph{Weights}: the released checkpoint versus our reproduction trained from the
released code (\emph{retrained}); \emph{Environment}: the paper's published
numbers versus our protocol. The \trianglemodel{} paper reports R@1 only, so the
published R@10 is omitted (--).}
\label{tab:supp-triangle-audit}
\scriptsize
\setlength{\tabcolsep}{2.6pt}
\renewcommand{\arraystretch}{1.05}
\begin{tabular}{@{}l cc cc cc c cc cc cc@{}}
\toprule
& \multicolumn{6}{c}{Text $\to$ Video} & & \multicolumn{6}{c}{Video $\to$ Text} \\
\cmidrule(lr){2-7}\cmidrule(lr){9-14}
\emph{Weights}
& \multicolumn{2}{c}{released} & \multicolumn{2}{c}{released} & \multicolumn{2}{c}{retrained}
& & \multicolumn{2}{c}{released} & \multicolumn{2}{c}{released} & \multicolumn{2}{c}{retrained} \\
\emph{Environment}
& \multicolumn{2}{c}{published} & \multicolumn{2}{c}{ours} & \multicolumn{2}{c}{ours}
& & \multicolumn{2}{c}{published} & \multicolumn{2}{c}{ours} & \multicolumn{2}{c}{ours} \\
\cmidrule(lr){2-3}\cmidrule(lr){4-5}\cmidrule(lr){6-7}\cmidrule(lr){9-10}\cmidrule(lr){11-12}\cmidrule(lr){13-14}
Dataset & R@1 & R@10 & R@1 & R@10 & R@1 & R@10 & & R@1 & R@10 & R@1 & R@10 & R@1 & R@10 \\
\midrule
MSR-VTT     & 55.2 & -- & 50.5 & 79.4 & 47.2 & 72.0 & & 52.5 & -- & 49.7 & 80.3 & 48.6 & 81.1 \\
DiDeMo      & 54.9 & -- & 51.0 & 75.5 & 48.4 & 70.6 & & 53.1 & -- & 49.4 & 79.2 & 50.7 & 79.5 \\
ActivityNet & 59.7 & -- & 56.0 & 83.8 & 51.3 & 75.0 & & 54.1 & -- & 52.3 & 87.2 & 52.7 & 87.7 \\
VATEX       & 83.9 & -- & 88.9 & 99.8 & 88.9 & 98.1 & & 80.9 & -- & 86.5 & 99.1 & 87.0 & 99.3 \\
\bottomrule
\end{tabular}
\end{table*}

\section{Limitations}
\label{sec:supp-limitations}
% -----------------------------------------------------------------------------
\paragraph{Backbone coverage.}
All experiments use the \vast{} backbone family, so that \gram{}, \hypergram{}, \pmrl{}, and \trianglemodel{} are compared under a shared initialisation, and \hyvol{} is instantiated on the two volume backbones
(\gram{} and \hypergram{}). Whether the same refinement gains hold for other
multimodal backbones remains to be evaluated.

\paragraph{Inference-inert regularization.}
The hypergraph module acts only during training: its hyperedges shape the
encoder weights, but the graph is removed at inference and the volume backbone
is scored on its own. The refinement is therefore fixed in the weights, cannot
be adapted per query or per corpus at test time, and its gain over the plain
volume backbone is modest and not uniform across benchmarks.

\paragraph{Continued-pretraining scale.}
Like the volume-based backbones it builds on, \hyvol{} performs continued
pretraining on a nominal 150K-clip subset of \vast{}-27M rather than the full corpus. The conclusions therefore apply to this continued-pretraining regime; their behaviour at substantially larger training-scale is not established.

\end{document}